\documentclass{article} 
\usepackage{iclr2026_conference,times}

\usepackage{amsmath,amsfonts,bm}

\def\eqref#1{equation~\ref{#1}}

\def\1{\bm{1}}

\DeclareMathAlphabet{\mathsfit}{\encodingdefault}{\sfdefault}{m}{sl}
\SetMathAlphabet{\mathsfit}{bold}{\encodingdefault}{\sfdefault}{bx}{n}

\usepackage{hyperref}
\usepackage{url}
\usepackage{xspace}
\usepackage{graphicx}
\usepackage{subcaption}
\usepackage{booktabs}
\usepackage{csquotes}
\usepackage{multirow}
\usepackage[table]{xcolor}

\newcommand{\methodname}{\textsc{CARPE}\xspace}

\title{CARPE: \\ Context-Aware Image Representation Prioritization via Ensemble for Large Vision-Language Models}

\author{Donghee Lee, Rui Cai, Zhe Zhao \\
Department of Computer Science\\
University of California, Davis\\
Davis, CA 95616, USA \\
\texttt{dhhlee@ucdavis.edu}
}

\iclrfinalcopy 
\begin{document}

\maketitle

\begin{abstract}
Large vision-language models (LVLMs) are typically trained using autoregressive language modeling objectives, which align visual representations with linguistic space. While effective for multimodal reasoning, this alignment can weaken vision-centric capabilities, causing LVLMs to underperform their base vision encoders on tasks such as image classification.
To address this limitation, we propose Context-Aware Image Representation Prioritization via Ensemble (\methodname), a lightweight framework that integrates raw vision features with aligned LLM representations through vision-integration layers and a context-aware ensemble mechanism.
This design enhances the model’s ability to adaptively weight visual and textual modalities and enables the model to capture various aspects of image representations. Extensive experiments demonstrate that \methodname improves performance on both image classification and diverse vision-language benchmarks. Our results suggest that modality balancing plays a critical role in multimodal generalization by improving representation utilization within autoregressive LVLMs.
\end{abstract}

\section{Introduction}
\label{introduction}
Large vision-language models (LVLMs) have become increasingly popular in the research community, as they serve as foundational building blocks towards general-purpose assistants \citep{LLaVA15_24, LLaVA_23, BLIP2_23, instructBLIP_23, minigpt4_23, Qwen2VL_24, mplugowl_23, internvl_24}. 
While existing LVLMs exhibit impressive performance across various vision-language tasks, recent studies have highlighted their limitations in image classification \citep{catastrophicforgettingmultimodal_23, classificationmultimodal_24, sparseattention_24}. Notably, \citet{catastrophicforgettingmultimodal_23} and \citet{classificationmultimodal_24} reveal that LVLMs significantly underperform CLIP \citep{CLIP_21} on standard image classification benchmarks such as ImageNet \citep{imagenet}, despite CLIP being their base vision encoder, indicating that LVLMs do not fully preserve the generalization properties of their underlying vision encoders.

\begin{figure}[t!]
    \includegraphics[width=0.5\textwidth]{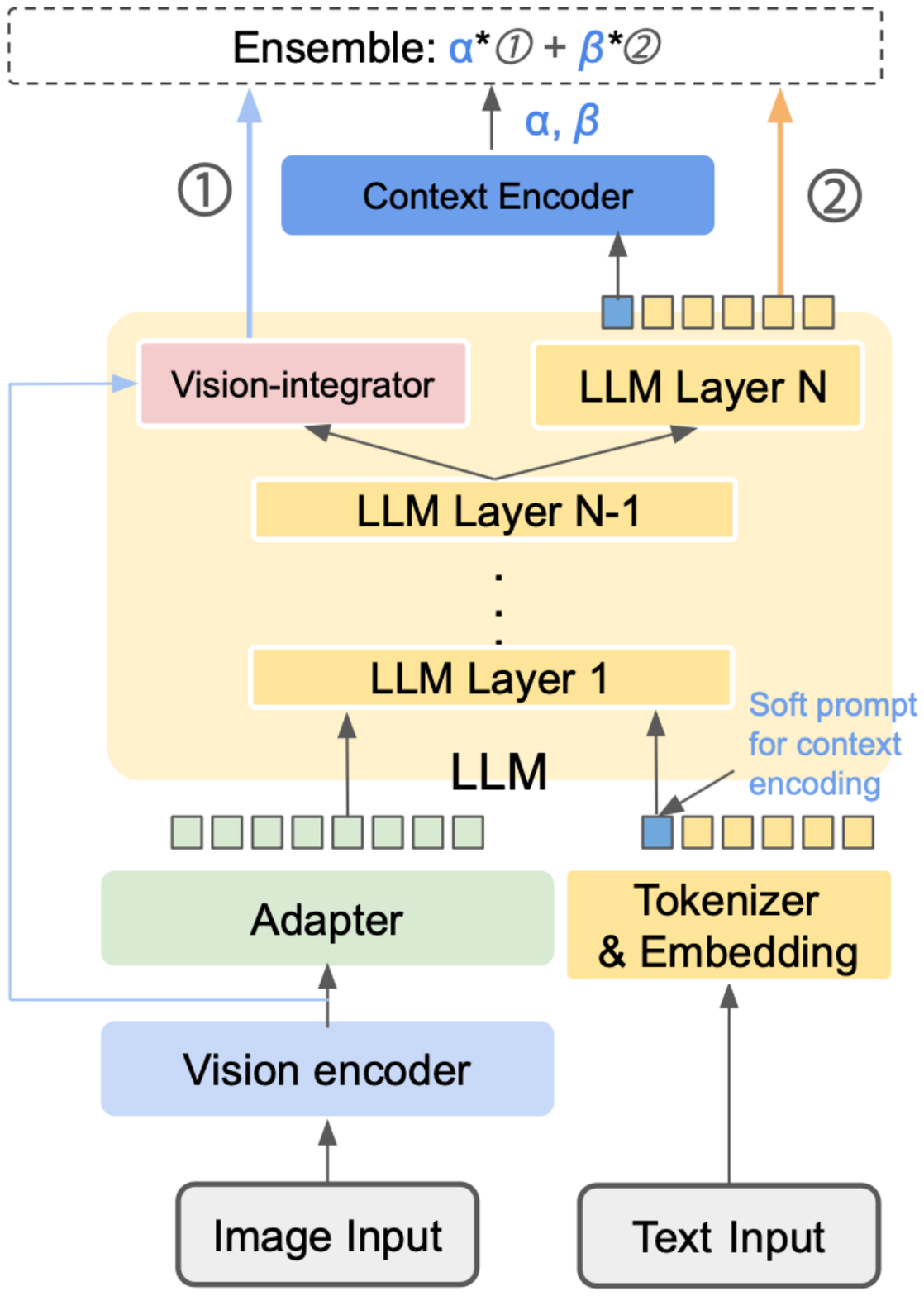}
    \small
    \centering
    \caption{\methodname architecture incorporates vision-integrator with a context-aware ensemble approach, dynamically combining vision representations from diverse perspectives.}
    \label{fig:framework}
\end{figure}

This underperformance in image classification presents a significant bottleneck for LVLMs. Although LVLMs are primarily designed for generative tasks, many vision-language tasks inherently rely on robust classification capabilities. The internal reasoning process often involves recognizing and categorizing visual elements before generating an answer. For instance, a sample from the TextVQA \citep{textvqa} benchmark presents the question, \enquote{\textit{What company made this?}} along with an image of a laptop. If the model fails to first classify the object as a laptop, subsequent reasoning steps are likely to be incorrect. Consequently, enhancing classification performance could naturally lead to broader improvements in LVLMs' overall capabilities.

\begin{table*}[t]
    \centering
    \small
    \renewcommand{\arraystretch}{1.2}
    \resizebox{\textwidth}{!}{
    \begin{tabular}{lccccccccc}
    \toprule
    \textbf{Model} & \multicolumn{4}{c}{\textbf{Classification}} & \multicolumn{5}{c}{\textbf{Vision-language Benchmarks}} \\
    \cmidrule(lr){2-5} \cmidrule(lr){6-10} 
    & \textbf{ImageNet} & \textbf{Caltech} & \textbf{Flowers102} & \textbf{Food101} & \textbf{SQA} & \textbf{MME} & \textbf{MMB} & \textbf{CV-Bench} & \textbf{MMVP} \\
    
    \midrule
    
    LLaVA1.5-7B \citep{LLaVA15_24} & 26.4 & 55.4 & 6.7 & 29.5 & 69.4 & 1862.7 & 64.7 & 46.0 & 63 \\ 
    + ImageNet fine-tune \citep{classificationmultimodal_24}  
    & 78 (+51.6) & 54.4 (-1.0) & 6.7 (-) & 22.9 (-6.6) & 66.8 (-2.6) & 1744.2 (-118.5) & 62.1 (-2.6) & 37.9 (-8.1) & 56.6 (-6.4) \\ 
    \bottomrule
    \end{tabular}
    }
    \caption{Performance of vanilla LLaVA1.5-7B and its ImageNet fine-tuned checkpoint on classification and vision-language benchmarks. The numbers in parentheses indicate the change in performance compared to the vanilla model.}
    \label{tab:imagenet_llava}
\end{table*}

 Building on the findings of \citet{classificationmultimodal_24}, we investigate whether fine-tuning LVLMs can enhance their general image classification performance. Our experiments demonstrate that while fine-tuning LLaVA1.5 \citep{LLaVA15_24} on ImageNet improves accuracy within the dataset, it simultaneously hurts the model's general capabilities. As shown in Table \ref{tab:imagenet_llava}, this fine-tuning approach results in decreased performance across multiple benchmarks. The declines in CV-Bench \citep{cambrian_24} and MMVP\citep{CLIP_blind_24} are particularly notable, as they are vision-centric benchmarks where classification was anticipated to provide benefits. This indicates that simply fine-tuning LVLMs on classification does not effectively generalize to other vision-language tasks, nor does it lead to consistent improvements in general visual understanding.

This observation leads us to investigate the following question: \textit{How can we enhance the general visual understanding of LVLMs by improving their classification ability?}

\begin{figure}[ht!]
    \centering
    \begin{subfigure}[b]{0.8\textwidth}
        \centering
        \includegraphics[width=\textwidth]{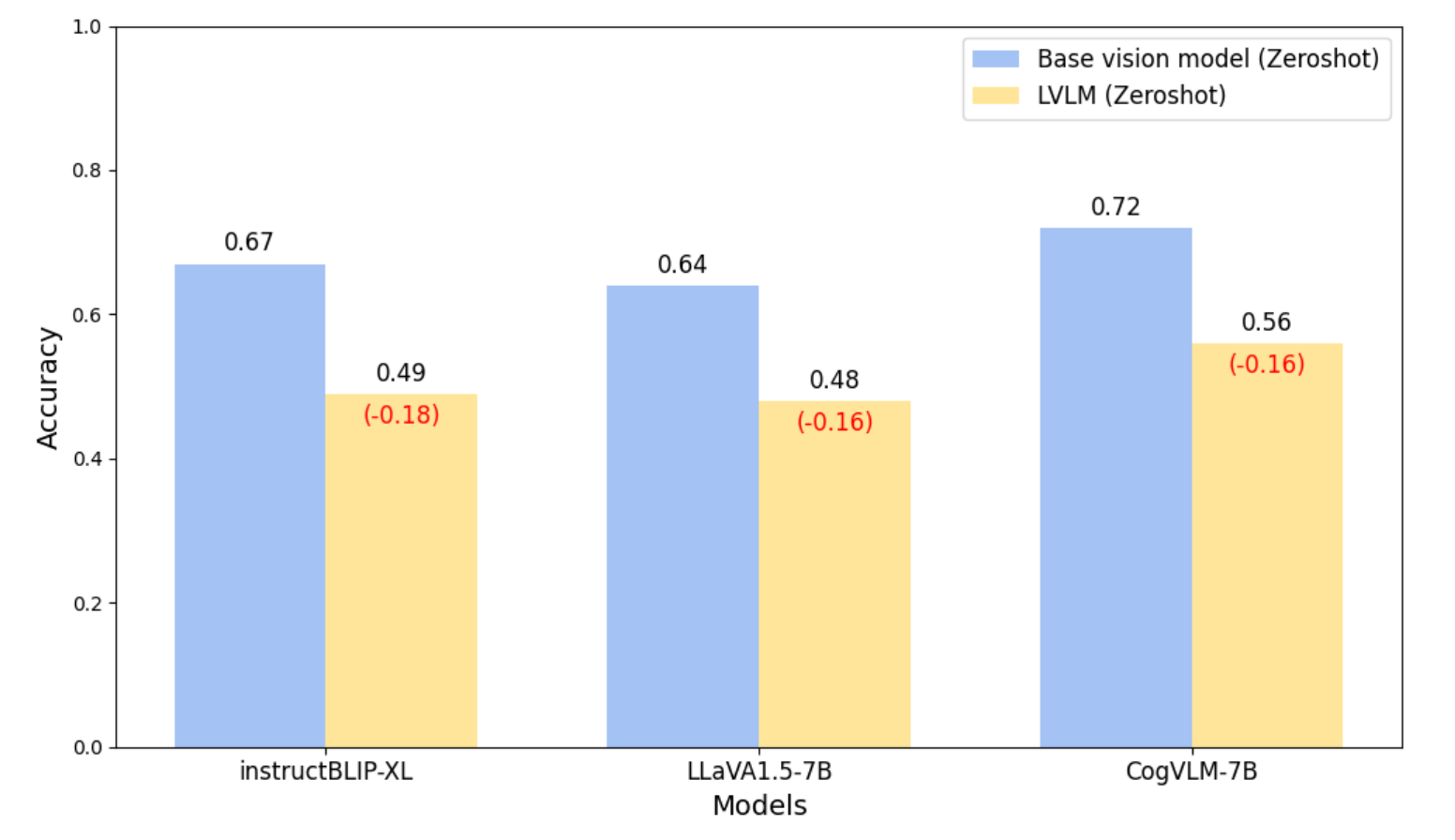}
        \caption{Zero-shot classification accuracy of base vision models (CLIP) and LVLMs.}
    \end{subfigure}
    \hfill
    \begin{subfigure}[b]{0.8\textwidth}
        \centering
        \includegraphics[width=\textwidth]{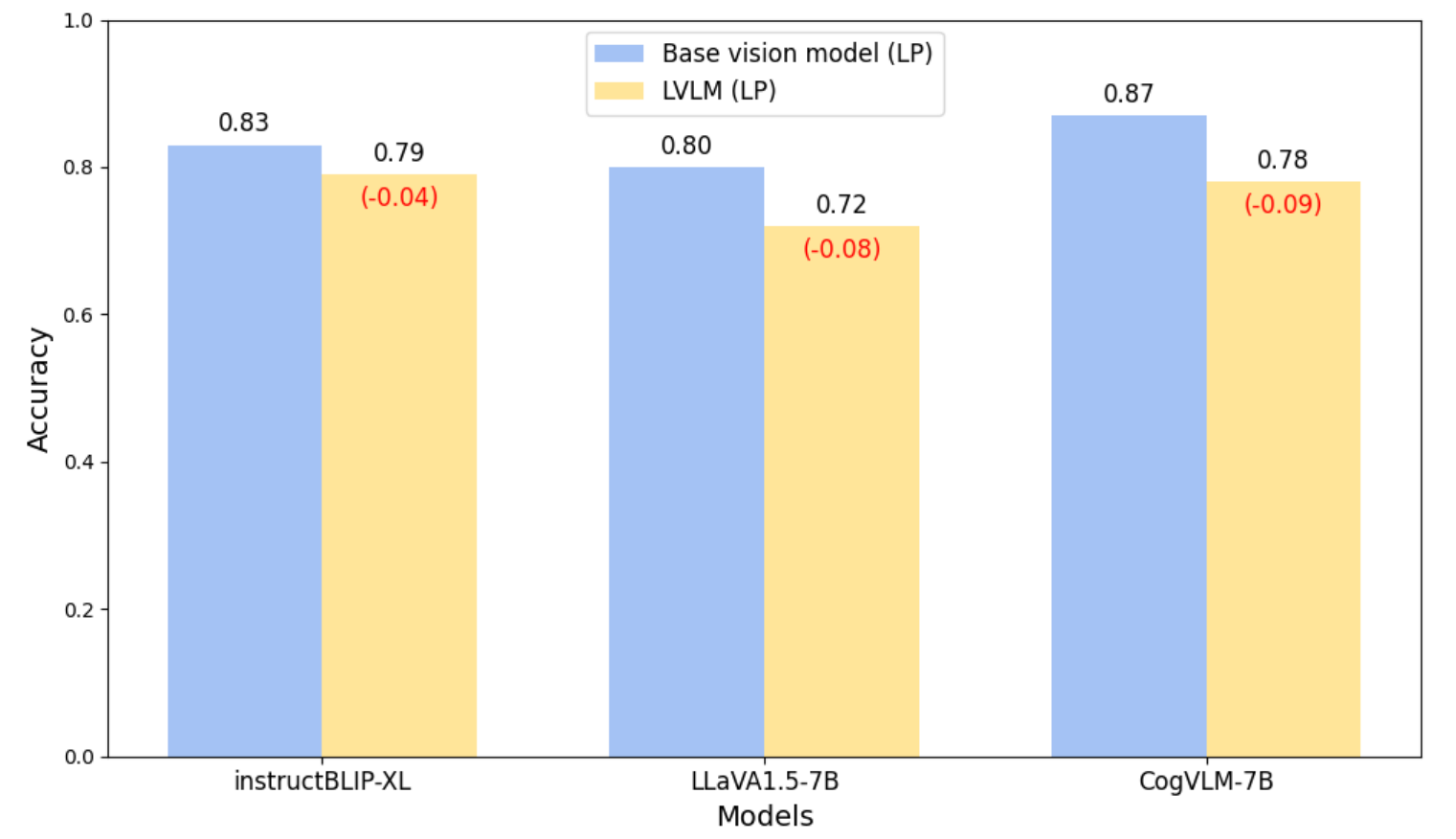}
        \caption{Linear probing classification accuracy comparing base vision models output and LVLMs final output.}
    \end{subfigure}
    \caption{Image classification performance analysis of LVLMs in zero-shot and linear probing settings. Accuracies are averaged across four classification datasets, Caltech101, Flowers102, DomainNet and Mini-ImageNet. }
    \label{fig:hypo_figure}
\end{figure}

Our key insight, supported by experimental results in Figure \ref{fig:hypo_figure}, is that classification-relevant information is largely retained within the LVLMs' latent space, despite being diminished in the final generated output. When evaluating three LVLMs on four classification datasets in a zero-shot setting, we observe a significant performance drop compared to their base CLIP models. However, when evaluated in a linear probing setting, they substantially close the performance gap with CLIP. These results indicate that classification-relevant visual information is largely retained in the LVLMs' latent space with minimal forgetting, as \citet{classificationmultimodal_24} found earlier. However, it becomes suboptimally aligned for downstream discriminative tasks during alignment with the LLM.

Based on this insight, we hypothesize that LVLMs struggle to adaptively discern when to prioritize image representations versus relying on language-based reasoning in a context-dependent manner. For example, in vision-centric tasks (e.g., image classification), which requires a strong focus on vision inputs, LVLMs may struggle to appropriately prioritize visual information over language model's reasoning capabilities due to the misalignment. This inability to balance visual and textual modalities based on the context could undermine their performance across diverse tasks.

To address this, we propose a dynamic ensemble approach that integrates two embeddings: (1) raw vision encodings directly from the vision encoder, and (2) final LLM outputs. These embeddings provide different visual perspectives, as prior studies have found that when vision features are aligned with language, they tend to focus on different aspects of information \citep{CLIP_21, PVIT_23, CLIP_blind_24, lion_24, pclip_22}. For example, \citet{CLIP_21} highlights that image-caption pairs emphasize high-level semantics over detailed descriptions. Therefore, by leveraging embeddings before and after LLM alignment, we aim to extract and utilize a richer set of visual features, enabling more effective utilization of suboptimally aligned visual representations.

Specifically, we fuse these complementary embeddings through a vision-integrator coupled with a context-aware ensemble mechanism. The vision-integrator captures different aspects of vision features, by combining vision encoder's output with LLM output to complement the visual information that may become misaligned during LLM's alignment process. Furthermore, the outputs of the vision-integrator and the final LLM output are combined via a context-aware ensemble mechanism guided by a context encoder. This context encoder dynamically adjusts ensemble weights based on context from text input, allowing the model to prioritize the vision-integrator output when pre-LLM alignment information is more relevant or the final LLM output when language-based reasoning is more required. 

Thus, we introduce this framework as \textbf{C}ontext-\textbf{A}ware Image \textbf{R}epresentation \textbf{P}rioritization via \textbf{E}nsemble (\methodname), which significantly improves performance on image classification benchmarks and extends benefits to various multimodal zero-shot tasks. Extensive experiments demonstrate that \methodname's context-aware ensemble design effectively integrates visual information from diverse perspectives to enhance their general capabilities. \methodname's framework can be seamlessly integrated with a wide range of open-source LVLMs that comprise a vision encoder and a language model comprising a vision encoder and a language model, ensuring broad applicability and ease of deployment.


Our contributions in this work are as follows:
\begin{itemize}
    \item We empirically analyze modality imbalance in LVLMs and show that classification-relevant information is largely preserved in latent representations but becomes underutilized after language alignment.
    
    \item We introduce \methodname, a novel framework that adaptively integrates multiple perspectives of visual features to enhance general image understanding in a context-dependent manner. 

    \item We demonstrate that our vision-integrator and context-aware ensemble improve generalization by enhancing classification performance.
    
\end{itemize}

\section{Related Works}
\label{related_works}
\paragraph{Large Vision-Language Models}
Recent LVLMs are predominantly built on pre-trained vision and large language models \citep{LLaVA_23, LLaVA15_24, instructBLIP_23, minigpt4_23, Qwen-VL_23, Qwen2VL_24, mplugowl_23, internvl_24}. These models are often connected using different types of adapter modules, such as MLPs \citep{LLaVA_23, LLaVA15_24, sphinx_23, sharegpt4v_23}, and Q-former \citep{BLIP2_23, instructBLIP_23} to integrate the different modalities. Models like Qwen2-VL \citep{Qwen2VL_24} and InternVL \citep{internvl_24} have showcased impressive capabilities in instruction-following and visual reasoning tasks, while some models are specifically designed to enhance visual understanding ability \citep{sphinx_23, cogvlm_23}. SPHINX \citep{sphinx_23} employs an ensemble of various vision backbones to extract robust visual representations from different aspects, and CogVLM \citep{cogvlm_23} introduces a visual expert module, doubling its parameters in language model to improve visual understanding abilities. In comparison, our work leverages light-weight vision-integrator to efficiently enhance image comprehension.

\paragraph{Limitations of LVLMs in image classification}
Although LVLMs showcase strong performance on many vision-language tasks, recent research underscores their shortcomings in image classification \citep{catastrophicforgettingmultimodal_23, classificationmultimodal_24, sparseattention_24, modalityinterference}. For example, \citet{catastrophicforgettingmultimodal_23} and \citet{classificationmultimodal_24} reveal that LVLMs fail to inherit the generalizability of CLIP on standard image classification tasks. 
\citet{modalityinterference} identify this failure as a cross-modality competency problem, where LVLMs struggle to fairly assess information across modalities.
In contrast, our framework introduces a dynamic weighting mechanism that adaptively prioritizes visual information, improving overall classification ability.

\paragraph{Vision features aligned to language overlook visual details}
Many studies have found that text-aligned image features emphasize high-level content while overlooking fine-grained details \citep{CLIP_21, PVIT_23, CLIP_blind_24, lion_24, pclip_22}. For instance, \citet{CLIP_21} suggests that image-caption pairs focus on high-level semantics description rather than visual details, leading to image representation primarily capture global features. 
To address this issue, \citet{pclip_22} proposes constructing three visual embeddings from different semantic levels for more accurate alignment between image and text in vision-language pre-training. In this work, we leverage the visual representation both before and after LLM alignment to effectively extract different aspect of visual information.

\section{Methods}
\subsection{Architecture}
In this section, we introduce \methodname's three key components: a pre-trained LVLM, vision-integrators and a context-aware weighting modules consisting of a context prompt and a context encoder. The overall framework of \methodname is summarized in Figure \ref{fig:framework}.

\paragraph{Pre-trained LVLM}
LVLMs typically consists of three parts: a vision encoder, which is commonly based on CLIP, a large language model (LLM), and an adapter that connects the two modalities, typically implemented as an MLP or a Q-former. Our framework is designed to be compatible with any LVLM that incorporates both vision and language components.

\paragraph{Vision-integrator}
We introduce a vision-integrator to effectively combine three types of visual information: (1) raw vision features from the base vision encoder (e.g., CLIP), (2) the LLM representations prior to the final vocabulary projection.

The motivation for this integration stems from previous studies indicating that aligning image features with language shifts the model's focus towards semantic level while losing fine-grained visual details. Based on this finding, we assume that each of these feature types \textemdash before and after LLM alignment \textemdash encodes complementary perspectives of the image. Thus, the vision-integrator is designed to efficiently merge these two forms of information, to enhance the model's comprehensive visual understanding. 

Specifically, vision-integrator consists of a multi-head cross-attention layer followed by multi-head self-attention layer and an MLP. The queries originate from the LLM's second-to-last output, while the keys and values are derived from the raw vision features. Since the raw vision features are not initially aligned with language space, they are first projected into the LLM's dimension using a newly introduced MLP adapter.

\paragraph{Context encoder and context prompt}
We introduce a context-aware weighting mechanism to enable the model to distinguish between vision- or language-centric contexts, rather than simply adding the two embeddings in an ensemble. We assume different tasks may require different weighting of these embeddings. 

To explicitly encode this ability, we incorporate a context prompt\textemdash a learnable soft prompt appended to the input text embeddings\textemdash motivated by prompt tuning \citep{prompt_tuning_21}. The context prompt is processed by the LLM as standard text input, and subsequently passed through the context encoder. The context encoder generates two probability values that sum to 1.0, which serve as ensemble weights between the vision-integrator and the final LLM representations.

We ensure that the context prompt is influenced solely by text inputs, as the distinction between vision-centric and language-centric tasks is determined by the instruction rather than the image content. In our implementation, the context encoder is a single linear layer followed by a softmax function, and the context prompt consists of a learnable embedding of length one. 
The final prediction is obtained by a weighted sum of the vision-integrator and LLM logits (see Appendix \ref{ensemble} for formulation).

\begin{figure}[ht]
    \centering
    \includegraphics[width=0.7\textwidth]{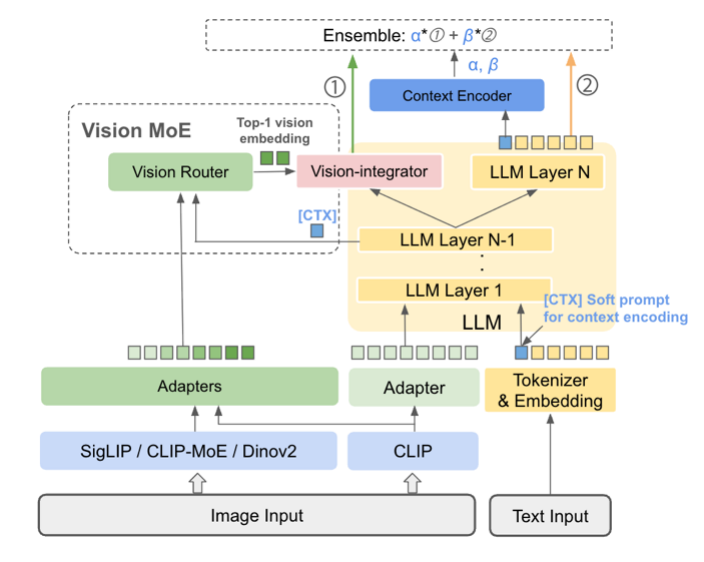}
    \caption{The architecture of \methodname-MoE. It extends the base \methodname model by incorporating a Vision MoE module that dynamically selects an expert from multiple vision backbones.}
    \label{fig:moe}
\end{figure}

\paragraph{Mixture of Experts}
Inspired by recent successes in Mixture-of-Experts (MoE) architectures \citep{mixtral, sphinx_23}, we introduce \methodname-MoE. As shown in Figure \ref{fig:moe}, this extension is designed to capture more robust visual representations through an ensemble of different vision encoders. 

Beyond the LVLM's base CLIP model, we add three pre-trained vision encoders as experts: SigLIP \citep{siglip}, DINOv2 \citep{dinov2}, and a CLIP-MoE model from the CuMo \citep{cumo} checkpoint. Each of the four backbones is paired with a dedicated two-layer MLP adapter, which projects its unique visual features into the LLM's common embedding space.

A linear vision router dynamically selects the most suitable expert for a given input. To maintain our framework's context-aware nature, the routing decision is conditioned on the hidden state of the learnable context prompt token. Based on this textual context, the router performs top-1 gating to direct the image to a single expert. Since each expert uses a lightweight two-layer MLP adapter and the vision router is implemented as a linear layer, \methodname-MoE introduces only a small number of additional trainable parameters compared to the base LVLM.

\section{Experiments}
\label{experiments}
\subsection{Experimental Setup}

We utilized LLaVA-Instruct-665K \citep{LLaVA15_24}, a publicly available instruction-tuning dataset, and combine it with Imagenet \citep{imagenet} to improve both classification ability and overall visual image comprehension. To avoid degrading language ability, we fix the mixing ratio at 1:7 (ImageNet: LLaVA-Instruct). For ImageNet prompting, we uniformly sample one of 20 classification prompt templates (i.e. ‘Identify the object in this image:’, ‘What object can you spot in the picture?’) per example. We mix open- and closed-world prompts 50/50 (without vs. with label lists) to reduce prompt overfitting and improve generalization.

In our experiments, we use LLaVA1.5-7b \citep{LLaVA15_24} as our base model to evaluate our framework. During training, we keep the adapter, final output projection head, vision-integrator, context encoder and context prompt trainable while freezing all remaining parts. We train the base \methodname model for 2 epochs and the \methodname-MoE model for 3 epochs, using batch size of 64 for all experiments. We set the learning rate to 2e-5 for the adapter and 2e-4 for the other trainable components. To stabilize the learning process, we freeze the context encoder and context prompt during the first epoch and unfreeze them in subsequent epochs.

\subsection{Baselines}
We compare CARPE with four baselines. As a classification fine-tuning baseline, we use the ImageNet-fine-tuned LLaVA-1.5-7B checkpoint \citep{classificationmultimodal_24}. We also include two ensemble baselines, WiSE-FT \citep{wiseft_21} and LEVI \citep{LEVI}. WiSE-FT \citep{wiseft_21} linearly interpolates the parameters of a zero-shot model and a fine-tuned model. In our setup, we mix the pre-trained LLaVA1.5-7B weights with the ImageNet-fine-tuned checkpoint \citep{classificationmultimodal_24} using a coefficient of 0.5. LEVI \citep{LEVI} adaptively ensembles a pre-trained model layer-wise with a small task-specific model to improve generalization in fine-tuning. To apply LEVI to generative LVLMs, we replace the task-specific branch with adapter outputs from the vision side and attach five adapting layers to the last five LLM layers. Each adapting layer performs multi-head cross-attention with queries from the corresponding LLM hidden states and keys and values from the adapter outputs, followed by multi-head self-attention and an MLP. The five adapted hidden states are averaged and projected onto the vocabulary space, producing final logits. Finally, we evaluate SPHINX \citep{sphinx_23} as a visually enhanced LVLM baseline that mixes model weights, training objectives, enriched visual embeddings, and high-resolution sub-images to improve overall capability. 


We evaluate \methodname on four image classification benchmarks\textemdash ImageNet, Caltech101, Flowers102, and Food101\textemdash and seven vision-language benchmarks, including SQA, TextVQA , POPE , MME , MMBench , CV-Bench , and MMVP (see Appendix \ref{datasets} for details).

\begin{table*}[h]
    \centering
    \small
    \renewcommand{\arraystretch}{1.2}
    \resizebox{\textwidth}{!}{
    \begin{tabular}{l l cccc c}
        \toprule
         & \textbf{Model} & ImageNet & Caltech101 & Flowers102 & Food101 & \textbf{Average} \\
        
        \midrule
        \multirow{5}{*}{Baseline} & LLaVA1.5-7B \citep{LLaVA15_24} & 26.4 & 55.4 & 6.7 & 29.5 & \cellcolor{gray!15}29.5 \\
        &  + Imagenet Fine-tune \citep{classificationmultimodal_24} & 78.0 & 54.4 & 6.7 & 22.9 & \cellcolor{gray!15}40.5 \\
        & WiSE-FT \citep{wiseft_21} & 48.1 & 56.2 & 6.6 & 26.9 & \cellcolor{gray!15}34.4 \\
        & LEVI \citep{LEVI} & 73.3 & 43.5 & 3.7 & 20.8 & \cellcolor{gray!15}35.3 \\
        & SPHINX-13B \citep{sphinx_23} & 32.3 & 49.7 & 17.7 & 36.3 & 
        \cellcolor{gray!15}34.0 \\
        \midrule
        \multirow{2}{*}{Ours} & \methodname & 73.4 & 60.4 & 15.4 & 32.7 & \cellcolor{gray!15}\underline{45.4} \\
        & \methodname-MoE & 64.5 & 65.6 & 16.7 & 37.7 & \cellcolor{gray!15}\textbf{46.1}\\
        \bottomrule
    \end{tabular}
    }
    \caption{Classification Accuracy (\%)}
    \label{tab:cls_results}
\end{table*}

\begin{table*}[h]
    \centering
    \small
    \renewcommand{\arraystretch}{1.2}
    \resizebox{\textwidth}{!}{
    \begin{tabular}{@{}l l ccccc cc c@{}}
        \toprule
        & \multirow{2}{*}{\textbf{Model}}
        & \multicolumn{5}{c}{General VL Benchmarks}
        & \multicolumn{2}{c}{Vision-centric VL}
        & \multirow{2}{*}{\textbf{Average}} \\

        \cmidrule(lr){3-7}\cmidrule(lr){8-9}
        & 
        & SQA & TextVQA & POPE & MME & MMBench 
        & CV-Bench & MMVP 
        & \\
        
        \midrule
        \multirow{5}{*}{Baseline} & LLaVA1.5-7B \citep{LLaVA15_24} & 69.4 & 58.3 & 85.9 & 1862.7 & 64.7 & 46.0 & 63.0 & \cellcolor{gray!15}68.6 \\
        &  + Imagenet Fine-tune \citep{classificationmultimodal_24} & 66.8 & 57.0 & 85.9 & 1744.2 & 62.1 & 37.9 & 56.6 & \cellcolor{gray!15}64.7 \\
        & WiSE-FT \citep{wiseft_21} & 68.0 & 57.8 & 85.1 & 1803.5 & 64.0 & 46.1 & 59.0 & \cellcolor{gray!15}67.1 \\
        & LEVI \citep{LEVI} & 69.4 & 49.2 & 84.5 & 1752.1 & 64.0 & 47.4 & 61.3 & \cellcolor{gray!15}66.2 \\
        & SPHINX-13B \citep{sphinx_23} & 69.3 & 51.6 & 80.7 & 1798.3 & 66.9 & 61.3 & 66.6 & \cellcolor{gray!15}69.4 \\
        
        \midrule
        \multirow{2}{*}{Ours} & \methodname & 68.4 & 55.8 & 85.2 & 1826.5 & 64.8 & 58.8 & 64.0 & \cellcolor{gray!15}\underline{69.7} \\
        & \methodname-MoE & 68.0 & 57.4 & 84.7 & 1861.7 & 64.0 & 58.8 & 65.0 & \cellcolor{gray!15}\textbf{70.1}\\
        \bottomrule
    \end{tabular}
    }
    \caption{Performance on seven vision-language benchmarks, including both general-purpose and vision-centric tasks. MME scores are scaled to 100 for averaging; SQA refers to the image subset of ScienceQA; POPE is reported with F1 score; all others are accuracy.}
    \label{tab:vqa_results}
\end{table*}

\section{Results}
\subsection{Classification Benchmarks}
As shown in Table \ref{tab:cls_results}, \methodname improves performance not only on the in-distribution dataset (ImageNet) but also on all out-of-distribution (OOD) classification benchmarks compared to the base LLaVA1.5-7B \citep{LLaVA15_24} model. The ImageNet fine-tuning baseline \citep{classificationmultimodal_24} yields a substantial gain on the in-distribution dataset, but it causes a notable performance drop on OOD datasets such as Caltech101 and Food101. In contrast, both \methodname and \methodname-MoE increase accuracy across both in-distribution and all OOD datasets. Notably, \methodname-MoE achieves the highest average classification accuracy, demonstrating the benefit of integrating diverse visual representations from multiple backbones for classification tasks.

When compared to other ensemble-based baselines such as WiSE-FT \citep{wiseft_21} and LEVI \citep{LEVI}, \methodname demonstrates clear superiority. Remarkably, despite SPHINX-13B \citep{sphinx_23} having nearly twice the model size, \methodname still surpasses it on three out of four benchmarks, highlighting the parameter efficiency of our design.

\subsection{Vision-Language Benchmarks}
Table \ref{tab:vqa_results} shows that both of our models outperform the baselines, with \methodname-MoE achieving the highest average performance across vision-language benchmarks. While ImageNet fine-tuning \citep{classificationmultimodal_24} leads to severe degradation in several benchmarks, \methodname and \methodname-MoE preserve performance on general benchmarks and deliver substantial improvements on vision-centric benchmarks. In particular, compared to the base model, \methodname-MoE improves CV-Bench and MMVP scores by 12.2\% and 2.0\%, respectively, whereas the fine-tuning baseline suffers large drops. Unlike LEVI, which relies on adapter outputs already projected into the language space, \methodname directly leverages raw vision features from the encoder, mitigating information loss during alignment.

These results support our hypothesis that visual features extracted from the vision encoder are partially distorted or lose generalizability during alignment with the LLM. By introducing the vision-integrator to incorporate raw vision features while retaining their original granularity and balancing them with LLM outputs in a context-dependent manner, \methodname enables more effective utilization of visual information that may be under-emphasized during LLM alignment.


\begin{table}[h]
    \centering
    \setlength{\abovecaptionskip}{6pt}
    \begin{tabular}{lcc}
        \toprule
        \textbf{Benchmark Type} & \textbf{Vision Weight} & \textbf{Language Weight} \\
        \midrule
        Classification          & 0.26 & 0.74 \\
        Vision-language Benchmark & 0.13 & 0.87 \\
        \bottomrule
    \end{tabular}
    \caption{Average vision and language weights of \methodname-MoE assigned by the context encoder across classification and vision-language benchmarks. 
    }
    \label{tab:context_weights}
\end{table}

\subsection{Context-Aware Balancing of Vision and Language}
Our design assumes that the optimal weighting between vision and language components depends on the task. Vision-centric tasks such as image classification should rely more heavily on raw vision features, while general vision-language tasks often require stronger utilization of the LLM’s reasoning ability. To enable this adaptability, we introduced a learnable context prompt and a lightweight context encoder to dynamically adjust the ensemble weights according to the text instruction.

Table \ref{tab:context_weights} shows that the vision-to-language weight ratio assigned by the context encoder differs between classification and vision-language benchmarks. Specifically, the average vision weight is 0.26 for classification benchmarks and 0.13 for vision-language benchmarks, indicating that the model allocates a relatively greater proportion of attention to vision features in classification tasks compared to vision-language tasks. This difference likely arises because classification tasks depend more heavily on precise visual recognition, whereas many vision-language benchmarks place greater emphasis on language-based reasoning and instruction-following. Such adaptive weighting reflects \methodname’s ability to adjust the balance between vision and language components according to task requirements, contributing to its strong performance across both classification and vision-language evaluations.

\section{Limitations}
While \methodname demonstrates improvements in image classification and several vision-language benchmarks, there are opportunities to improve this work. Our experiments are conducted using a single representative LVLM, LLaVA-1.5, and we intentionally adopt a simple design for the vision-integrator and context encoder, where the vision-integrator consists of a lightweight cross-attention module composed of cross-attention, self-attention, and an MLP, and the context encoder is implemented as a single linear layer. Although this design keeps \methodname parameter-efficient and easy to integrate into existing LVLMs, further improvements may be possible by exploring richer architectural choices, such as varying the number of attention layers or heads, using different LLM layers for feature extraction, or evaluating the framework with more recent base LVLMs. In addition, while \methodname improves average performance, it shows performance drops on several general vision-language benchmarks, and its classification accuracy, although improved over the base LVLM, still remains below that of the base CLIP model. Future work will explore alternative ensemble strategies, architectural variations, and evaluations on stronger LVLM architectures to further improve performance and generalizability.

\section{Conclusion}
In this work, we addressed the challenge of enhancing the general visual understanding of LVLMs by improving their classification capability. We proposed \methodname, a lightweight vision-integration module combined with a context-aware dynamic ensemble strategy. Furthermore, we introduced \methodname-MoE, an extension that incorporates a Mixture of Experts framework to leverage multiple, diverse vision backbones.

Extensive experiments demonstrate that our methods effectively recover visual information that may be lost during LLM alignment, leading to performance gains on both classification and vision-language benchmarks. In particular, the \methodname-MoE variant demonstrate superior generalization, achieving the highest average performance across all evaluated tasks. These results confirm that improving general visual understanding also benefits broader vision-language capabilities. Finally, we showed that different tasks require different weighting of vision and language components, and by adaptively balancing the two based on context, our approach enhances the overall multimodal reasoning ability of LVLMs.

\section*{Ethics statement}
This work adheres to the ICLR Code of Ethics\footnote{\url{https://iclr.cc/public/CodeOfEthics}}. Our research focuses on analyzing and mitigating modality imbalance in large vision-language models through a context-aware integration framework. We aim to conduct and report our research transparently, minimize potential harm, and consider the broader societal implications of multimodal systems. However, as with other LLM research, we acknowledge that there remains potential risk that such systems may generate biased or harmful outputs embedded in pretrained models. All datasets used in our experiments (e.g., ImageNet, Caltech101, Flowers102, Food101, and standard vision-language benchmarks) are publicly available and widely adopted in the research community.

\section*{Reproducibility Statement}
We designed our experiments to be fully reproducible. Our implementation builds on the publicly available LLaVA codebase\footnote{\url{https://github.com/haotian-liu/LLaVA}}, and all experiments are conducted using publicly available pretrained vision-language models and widely adopted benchmark datasets. Experimental settings, implementation details, and hyperparameters are specified in Section \ref{experiments} and the appendix. The complete implementation of \methodname, including training scripts, evaluation pipelines, and configuration files, is available at \url{https://github.com/donghee1ee/CARPE}.

\bibliography{references}

@String(CVPR= {IEEE Conf. Comput. Vis. Pattern Recog.})

@String(CVPR  = {CVPR})

@InProceedings{LLaVA15_24,
    author    = {Liu, Haotian and Li, Chunyuan and Li, Yuheng and Lee, Yong Jae},
    title     = {Improved Baselines with Visual Instruction Tuning},
    booktitle = {Proceedings of the IEEE/CVF Conference on Computer Vision and Pattern Recognition (CVPR)},
    month     = {June},
    year      = {2024},
    pages     = {26296-26306}
}

@inproceedings{LLaVA_23,
 author = {Liu, Haotian and Li, Chunyuan and Wu, Qingyang and Lee, Yong Jae},
 booktitle = {Advances in Neural Information Processing Systems},
 editor = {A. Oh and T. Naumann and A. Globerson and K. Saenko and M. Hardt and S. Levine},
 pages = {34892--34916},
 publisher = {Curran Associates, Inc.},
 title = {Visual Instruction Tuning},
 url = {https://proceedings.neurips.cc/paper_files/paper/2023/file/6dcf277ea32ce3288914faf369fe6de0-Paper-Conference.pdf},
 volume = {36},
 year = {2023}
}

@inproceedings{BLIP2_23,
      title={{BLIP-2:} Bootstrapping Language-Image Pre-training with Frozen Image Encoders and Large Language Models}, 
      author={Junnan Li and Dongxu Li and Silvio Savarese and Steven Hoi},
      year={2023},
      booktitle={ICML},
}

@inproceedings{instructBLIP_23,
 author = {Dai, Wenliang and Li, Junnan and LI, DONGXU and Tiong, Anthony and Zhao, Junqi and Wang, Weisheng and Li, Boyang and Fung, Pascale N and Hoi, Steven},
 booktitle = {Advances in Neural Information Processing Systems},
 editor = {A. Oh and T. Naumann and A. Globerson and K. Saenko and M. Hardt and S. Levine},
 pages = {49250--49267},
 publisher = {Curran Associates, Inc.},
 title = {InstructBLIP: Towards General-purpose Vision-Language Models with Instruction Tuning},
 url = {https://proceedings.neurips.cc/paper_files/paper/2023/file/9a6a435e75419a836fe47ab6793623e6-Paper-Conference.pdf},
 volume = {36},
 year = {2023}
}

@article{minigpt4_23,
  title={MiniGPT-4: Enhancing Vision-Language Understanding with Advanced Large Language Models},
  author={Zhu, Deyao and Chen, Jun and Shen, Xiaoqian and Li, Xiang and Elhoseiny, Mohamed},
  journal={arXiv preprint arXiv:2304.10592},
  year={2023}
}

@article{Qwen-VL_23,
  title={Qwen-VL: A Versatile Vision-Language Model for Understanding, Localization, Text Reading, and Beyond},
  author={Bai, Jinze and Bai, Shuai and Yang, Shusheng and Wang, Shijie and Tan, Sinan and Wang, Peng and Lin, Junyang and Zhou, Chang and Zhou, Jingren},
  journal={arXiv preprint arXiv:2308.12966},
  year={2023}
}

@article{Qwen2VL_24,
  title={Qwen2-VL: Enhancing Vision-Language Model's Perception of the World at Any Resolution},
  author={Wang, Peng and Bai, Shuai and Tan, Sinan and Wang, Shijie and Fan, Zhihao and Bai, Jinze and Chen, Keqin and Liu, Xuejing and Wang, Jialin and Ge, Wenbin and Fan, Yang and Dang, Kai and Du, Mengfei and Ren, Xuancheng and Men, Rui and Liu, Dayiheng and Zhou, Chang and Zhou, Jingren and Lin, Junyang},
  journal={arXiv preprint arXiv:2409.12191},
  year={2024}
}

@misc{mplugowl_23,
      title={mPLUG-Owl: Modularization Empowers Large Language Models with Multimodality}, 
      author={Qinghao Ye and Haiyang Xu and Guohai Xu and Jiabo Ye and Ming Yan and Yiyang Zhou and Junyang Wang and Anwen Hu and Pengcheng Shi and Yaya Shi and Chaoya Jiang and Chenliang Li and Yuanhong Xu and Hehong Chen and Junfeng Tian and Qi Qian and Ji Zhang and Fei Huang},
      year={2023},
      eprint={2304.14178},
      archivePrefix={arXiv},
      primaryClass={cs.CL}
}

@inproceedings{internvl_24,
  title={Internvl: Scaling up vision foundation models and aligning for generic visual-linguistic tasks},
  author={Chen, Zhe and Wu, Jiannan and Wang, Wenhai and Su, Weijie and Chen, Guo and Xing, Sen and Zhong, Muyan and Zhang, Qinglong and Zhu, Xizhou and Lu, Lewei and others},
  booktitle={Proceedings of the IEEE/CVF Conference on Computer Vision and Pattern Recognition},
  pages={24185--24198},
  year={2024}
}

@article{sharegpt4v_23,
    title={ShareGPT4V: Improving Large Multi-Modal Models with Better Captions},
    author={Chen, Lin and Li, Jisong and Dong, Xiaoyi and Zhang, Pan and He, Conghui and Wang, Jiaqi and Zhao, Feng and Lin, Dahua},
    journal={arXiv preprint arXiv:2311.12793},
    year={2023}
}

@misc{sphinx_23,
      title={SPHINX: The Joint Mixing of Weights, Tasks, and Visual Embeddings for Multi-modal Large Language Models}, 
      author={Ziyi Lin and Chris Liu and Renrui Zhang and Peng Gao and Longtian Qiu and Han Xiao and Han Qiu and Chen Lin and Wenqi Shao and Keqin Chen and Jiaming Han and Siyuan Huang and Yichi Zhang and Xuming He and Hongsheng Li and Yu Qiao},
      year={2023},
      eprint={2311.07575},
      archivePrefix={arXiv},
      primaryClass={cs.CV},
      url={https://arxiv.org/abs/2311.07575}, 
}

@misc{catastrophicforgettingmultimodal_23,
      title={Investigating the Catastrophic Forgetting in Multimodal Large Language Models}, 
      author={Yuexiang Zhai and Shengbang Tong and Xiao Li and Mu Cai and Qing Qu and Yong Jae Lee and Yi Ma},
      year={2023},
      eprint={2309.10313},
      archivePrefix={arXiv},
      primaryClass={cs.CL},
      url={https://arxiv.org/abs/2309.10313}, 
}

@inproceedings{
classificationmultimodal_24,
title={Why are Visually-Grounded Language Models Bad at Image Classification?},
author={Yuhui Zhang and Alyssa Unell and Xiaohan Wang and Dhruba Ghosh and Yuchang Su and Ludwig Schmidt and Serena Yeung-Levy},
booktitle={The Thirty-eighth Annual Conference on Neural Information Processing Systems},
year={2024},
url={https://openreview.net/forum?id=MwmmBg1VYg}
}

@misc{sparseattention_24,
      title={Sparse Attention Vectors: Generative Multimodal Model Features Are Discriminative Vision-Language Classifiers}, 
      author={Chancharik Mitra and Brandon Huang and Tianning Chai and Zhiqiu Lin and Assaf Arbelle and Rogerio Feris and Leonid Karlinsky and Trevor Darrell and Deva Ramanan and Roei Herzig},
      year={2024},
      eprint={2412.00142},
      archivePrefix={arXiv},
      primaryClass={cs.CV},
      url={https://arxiv.org/abs/2412.00142}, 
}

@InProceedings{CLIP_21,
  title = 	 {Learning Transferable Visual Models From Natural Language Supervision},
  author =       {Radford, Alec and Kim, Jong Wook and Hallacy, Chris and Ramesh, Aditya and Goh, Gabriel and Agarwal, Sandhini and Sastry, Girish and Askell, Amanda and Mishkin, Pamela and Clark, Jack and Krueger, Gretchen and Sutskever, Ilya},
  booktitle = 	 {Proceedings of the 38th International Conference on Machine Learning},
  pages = 	 {8748--8763},
  year = 	 {2021},
  editor = 	 {Meila, Marina and Zhang, Tong},
  volume = 	 {139},
  series = 	 {Proceedings of Machine Learning Research},
  month = 	 {18--24 Jul},
  publisher =    {PMLR},
  url = 	 {https://proceedings.mlr.press/v139/radford21a.html},
}

@INPROCEEDINGS{imagenet,
  author={Deng, Jia and Dong, Wei and Socher, Richard and Li, Li-Jia and Kai Li and Li Fei-Fei},
  booktitle={2009 IEEE Conference on Computer Vision and Pattern Recognition}, 
  title={ImageNet: A large-scale hierarchical image database}, 
  year={2009},
  volume={},
  number={},
  pages={248-255},
  doi={10.1109/CVPR.2009.5206848}}

@InProceedings{LEVI,
  title = 	 {{LEVI}: Generalizable Fine-tuning via Layer-wise Ensemble of Different Views},
  author =       {Roh, Yuji and Liu, Qingyun and Gui, Huan and Yuan, Zhe and Tang, Yujin and Whang, Steven Euijong and Liu, Liang and Bi, Shuchao and Hong, Lichan and Chi, Ed H. and Zhao, Zhe},
  booktitle = 	 {Proceedings of the 41st International Conference on Machine Learning},
  pages = 	 {42666--42690},
  year = 	 {2024},
  editor = 	 {Salakhutdinov, Ruslan and Kolter, Zico and Heller, Katherine and Weller, Adrian and Oliver, Nuria and Scarlett, Jonathan and Berkenkamp, Felix},
  volume = 	 {235},
  series = 	 {Proceedings of Machine Learning Research},
  month = 	 {21--27 Jul},
  publisher =    {PMLR},
  url = 	 {https://proceedings.mlr.press/v235/roh24a.html},
}

@misc{cogvlm_23,
      title={CogVLM: Visual Expert for Pretrained Language Models}, 
      author={Weihan Wang and Qingsong Lv and Wenmeng Yu and Wenyi Hong and Ji Qi and Yan Wang and Junhui Ji and Zhuoyi Yang and Lei Zhao and Xixuan Song and Jiazheng Xu and Bin Xu and Juanzi Li and Yuxiao Dong and Ming Ding and Jie Tang},
      year={2023},
      eprint={2311.03079},
      archivePrefix={arXiv},
      primaryClass={cs.CV}
}

@misc{PVIT_23,
      title={Position-Enhanced Visual Instruction Tuning for Multimodal Large Language Models}, 
      author={Chi Chen and Ruoyu Qin and Fuwen Luo and Xiaoyue Mi and Peng Li and Maosong Sun and Yang Liu},
      year={2023},
      eprint={2308.13437},
      archivePrefix={arXiv},
      primaryClass={cs.CV}
}

@inproceedings{CLIP_blind_24,
  title={Eyes wide shut? exploring the visual shortcomings of multimodal llms},
  author={Tong, Shengbang and Liu, Zhuang and Zhai, Yuexiang and Ma, Yi and LeCun, Yann and Xie, Saining},
  booktitle={Proceedings of the IEEE/CVF Conference on Computer Vision and Pattern Recognition},
  pages={9568--9578},
  year={2024}
}

@inproceedings{lion_24,
  title={Lion: Empowering multimodal large language model with dual-level visual knowledge},
  author={Chen, Gongwei and Shen, Leyang and Shao, Rui and Deng, Xiang and Nie, Liqiang},
  booktitle={Proceedings of the IEEE/CVF Conference on Computer Vision and Pattern Recognition},
  pages={26540--26550},
  year={2024}
}

@misc{pclip_22,
      title={PyramidCLIP: Hierarchical Feature Alignment for Vision-language Model Pretraining}, 
      author={Yuting Gao and Jinfeng Liu and Zihan Xu and Jun Zhang and Ke Li and Rongrong Ji and Chunhua Shen},
      year={2022},
      eprint={2204.14095},
      archivePrefix={arXiv},
      primaryClass={cs.CV},
      url={https://arxiv.org/abs/2204.14095}, 
}

@misc{prompt_tuning_21,
      title={The Power of Scale for Parameter-Efficient Prompt Tuning}, 
      author={Brian Lester and Rami Al-Rfou and Noah Constant},
      year={2021},
      eprint={2104.08691},
      archivePrefix={arXiv},
      primaryClass={cs.CL},
      url={https://arxiv.org/abs/2104.08691}, 
}

@misc{wiseft_21,
      title={Robust fine-tuning of zero-shot models}, 
      author={Mitchell Wortsman and Gabriel Ilharco and Jong Wook Kim and Mike Li and Simon Kornblith and Rebecca Roelofs and Raphael Gontijo-Lopes and Hannaneh Hajishirzi and Ali Farhadi and Hongseok Namkoong and Ludwig Schmidt},
      year={2022},
      eprint={2109.01903},
      archivePrefix={arXiv},
      primaryClass={cs.CV},
      url={https://arxiv.org/abs/2109.01903}, 
}

@InProceedings{flowers102,
   author = "Nilsback, M-E. and Zisserman, A.",
   title = "Automated Flower Classification over a Large Number of Classes",
   booktitle = "Proceedings of the Indian Conference on Computer Vision, Graphics and Image Processing",
   year = "2008",
   month = "Dec"
}

@article{caltech101,
  title={Learning Generative Visual Models from Few Training Examples: An Incremental Bayesian Approach Tested on 101 Object Categories},
  author={Li Fei-Fei and Rob Fergus and Pietro Perona},
  journal={Computer Vision and Pattern Recognition Workshop},
  year={2004},
}

@misc{scienceqa,
      title={Learn to Explain: Multimodal Reasoning via Thought Chains for Science Question Answering}, 
      author={Pan Lu and Swaroop Mishra and Tony Xia and Liang Qiu and Kai-Wei Chang and Song-Chun Zhu and Oyvind Tafjord and Peter Clark and Ashwin Kalyan},
      year={2022},
      eprint={2209.09513},
      archivePrefix={arXiv},
      primaryClass={cs.CL},
      url={https://arxiv.org/abs/2209.09513}, 
}

@misc{textvqa,
      title={Towards VQA Models That Can Read}, 
      author={Amanpreet Singh and Vivek Natarajan and Meet Shah and Yu Jiang and Xinlei Chen and Dhruv Batra and Devi Parikh and Marcus Rohrbach},
      year={2019},
      eprint={1904.08920},
      archivePrefix={arXiv},
      primaryClass={cs.CL},
      url={https://arxiv.org/abs/1904.08920}, 
}

@misc{pope,
      title={Evaluating Object Hallucination in Large Vision-Language Models}, 
      author={Yifan Li and Yifan Du and Kun Zhou and Jinpeng Wang and Wayne Xin Zhao and Ji-Rong Wen},
      year={2023},
      eprint={2305.10355},
      archivePrefix={arXiv},
      primaryClass={cs.CV},
      url={https://arxiv.org/abs/2305.10355}, 
}

@misc{mme,
      title={MME: A Comprehensive Evaluation Benchmark for Multimodal Large Language Models}, 
      author={Chaoyou Fu and Peixian Chen and Yunhang Shen and Yulei Qin and Mengdan Zhang and Xu Lin and Jinrui Yang and Xiawu Zheng and Ke Li and Xing Sun and Yunsheng Wu and Rongrong Ji},
      year={2024},
      eprint={2306.13394},
      archivePrefix={arXiv},
      primaryClass={cs.CV},
      url={https://arxiv.org/abs/2306.13394}, 
}

@misc{mmbench,
      title={MMBench: Is Your Multi-modal Model an All-around Player?}, 
      author={Yuan Liu and Haodong Duan and Yuanhan Zhang and Bo Li and Songyang Zhang and Wangbo Zhao and Yike Yuan and Jiaqi Wang and Conghui He and Ziwei Liu and Kai Chen and Dahua Lin},
      year={2024},
      eprint={2307.06281},
      archivePrefix={arXiv},
      primaryClass={cs.CV},
      url={https://arxiv.org/abs/2307.06281}, 
}

@misc{cambrian_24,
      title={Cambrian-1: A Fully Open, Vision-Centric Exploration of Multimodal LLMs}, 
      author={Shengbang Tong and Ellis Brown and Penghao Wu and Sanghyun Woo and Manoj Middepogu and Sai Charitha Akula and Jihan Yang and Shusheng Yang and Adithya Iyer and Xichen Pan and Austin Wang and Rob Fergus and Yann LeCun and Saining Xie},
      year={2024},
      eprint={2406.16860},
}

@inproceedings{food101,
  title = {Food-101 -- Mining Discriminative Components with Random Forests},
  author = {Bossard, Lukas and Guillaumin, Matthieu and Van Gool, Luc},
  booktitle = {European Conference on Computer Vision},
  year = {2014}
}

@misc{mixtral,
      title={Mixtral of Experts}, 
      author={Albert Q. Jiang and Alexandre Sablayrolles and Antoine Roux and Arthur Mensch and Blanche Savary and Chris Bamford and Devendra Singh Chaplot and Diego de las Casas and Emma Bou Hanna and Florian Bressand and Gianna Lengyel and Guillaume Bour and Guillaume Lample and Lélio Renard Lavaud and Lucile Saulnier and Marie-Anne Lachaux and Pierre Stock and Sandeep Subramanian and Sophia Yang and Szymon Antoniak and Teven Le Scao and Théophile Gervet and Thibaut Lavril and Thomas Wang and Timothée Lacroix and William El Sayed},
      year={2024},
      eprint={2401.04088},
      archivePrefix={arXiv},
      primaryClass={cs.LG},
      url={https://arxiv.org/abs/2401.04088}, 
}

@misc{siglip,
      title={Sigmoid Loss for Language Image Pre-Training}, 
      author={Xiaohua Zhai and Basil Mustafa and Alexander Kolesnikov and Lucas Beyer},
      year={2023},
      eprint={2303.15343},
      archivePrefix={arXiv},
      primaryClass={cs.CV},
      url={https://arxiv.org/abs/2303.15343}, 
}

@misc{dinov2,
  title={DINOv2: Learning Robust Visual Features without Supervision},
  author={Oquab, Maxime and Darcet, Timothée and Moutakanni, Theo and Vo, Huy V. and Szafraniec, Marc and Khalidov, Vasil and Fernandez, Pierre and Haziza, Daniel and Massa, Francisco and El-Nouby, Alaaeldin and Howes, Russell and Huang, Po-Yao and Xu, Hu and Sharma, Vasu and Li, Shang-Wen and Galuba, Wojciech and Rabbat, Mike and Assran, Mido and Ballas, Nicolas and Synnaeve, Gabriel and Misra, Ishan and Jegou, Herve and Mairal, Julien and Labatut, Patrick and Joulin, Armand and Bojanowski, Piotr},
  journal={arXiv:2304.07193},
  year={2023}
}

@misc{cumo,
      title={CuMo: Scaling Multimodal LLM with Co-Upcycled Mixture-of-Experts}, 
      author={Jiachen Li and Xinyao Wang and Sijie Zhu and Chia-Wen Kuo and Lu Xu and Fan Chen and Jitesh Jain and Humphrey Shi and Longyin Wen},
      year={2024},
      eprint={2405.05949},
      archivePrefix={arXiv},
      primaryClass={cs.CV},
      url={https://arxiv.org/abs/2405.05949}, 
}

@article{modalityinterference,
  title={Diagnosing and Mitigating Modality Interference in Multimodal Large Language Models},
  author={Cai, Rui and Li, Bangzheng and Wen, Xiaofei and Chen, Muhao and Zhao, Zhe},
  journal={arXiv preprint arXiv:2505.19616},
  year={2025}
}
\bibliographystyle{iclr2026_conference}

\appendix
\section{Appendix}

\subsection{Using ensemble}
\label{ensemble}
To effectively combine the embeddings from the vision-integrator and the LLM representations, we employ an ensemble strategy that integrates their logits.

Formally, Let $X_{txt}$ be the input text sequence, $X_{img}$ the input image, and $Y$ a target token. Let $H_{vision}, H_{llm} \in \mathbb{R}^{N \times d}$ denote the hidden representations obtained from the vision integrator and the final LLM layer, where $N$ is the sequence length and $d$ is the hidden dimension. The shared output projection to the vocabulary is denoted by $W_{head} \in \mathbb{R}^{V \times d}$, where $V$ is the vocabulary size.

We first compute the logits from each representation as follows:
\[
Z_{vision} = H_{vision}W^T_{head}
\]
\[
Z_{llm} = H_{llm}W^T_{head}
\]

To determine context-aware ensemble weights, we introduce a learnable context prompt token, which is appended to the input and processed by the LLM. Let $H_{context} \in \mathbb{R}^{d}$ denote the final hidden state of the context prompt token, and let $W_{context} \in \mathbb{R}^{2 \times d}$ be the context encoder that projects this hidden state to a two-dimensional weight vector.

The ensemble weights are computed as:
\[
\alpha, \beta = \text{Softmax}(W_{context} H^T_{context})
\]

Using these weights, the final logit is computed as a weighted sum:
\[
Z = \alpha \cdot Z_{vision} + \beta \cdot Z_{llm}
\]

\subsection{Evaluation datasets}
\label{datasets}
To validate the effectiveness of \methodname, we evaluated the models on four classification datasets and seven vision-language benchmarks. The classification datasets include ImageNet \citep{imagenet}, Caltech101 \citep{caltech101}, Flower102 \citep{flowers102}, and Food101 \citep{food101}. Flowers102 \citep{flowers102} and Food101 \citep{food101} comprise 102 and 101 categories, respectively, and are used to evaluate fine-grained visual understanding. Caltech101 \citep{caltech101}, comprises 101 object classes and serves as a benchmark for assessing classification performance at a semantic level.

The vision-language benchmarks includes two academic-task-oriented datasets: image subset of ScienceQA \citep{scienceqa} and TextVQA \citep{textvqa}, which evaluates zero-shot generalization on scientific question answering and text-rich visual question answering, respectively. Benchmarks for instruction-following LVLMs include five datasets: POPE \citep{pope}, MME \citep{mme}, MMBench \citep{mmbench} CV-Bench \citep{cambrian_24}, and MMVP \citep{CLIP_blind_24}. These benchmarks assess various LVLMs abilities, including object hallucination, OCR perception, language generation, mathematical reasoning, scene understanding, and object counting. In particular, CV-Bench and MMVP are vision-centric benchmarks. CV-Bench \citep{cambrian_24}, evaluates classic vision tasks in multimodal settings, including spatial relations and depth ordering. MMVP \citep{CLIP_blind_24} targets CLIP-blind pairs, where CLIP judges visually distinct images as similar.

\end{document}